\documentclass[conference, onecolumn]{IEEEtran}
\IEEEoverridecommandlockouts
\usepackage{cite}
\usepackage{amsmath,amssymb,amsfonts}
\usepackage{algorithmic}
\usepackage{graphicx}
\usepackage{textcomp}
\usepackage[table]{xcolor}

\usepackage{url}

\usepackage{breakurl}

\usepackage{enumitem}
\usepackage{standalone}
\usepackage{tikz,pgfplots}
\usepackage{balance}

\usepackage{multicol}

	{\end{list}}

\usetikzlibrary{arrows.meta} 
\tikzset{>={Latex[length=4,width=4]}} 
\usetikzlibrary{calc}
\usepackage{amsmath,bm}
\usepackage{relsize} 

\usepackage[utf8]{inputenc}
\usepackage[polish,english]{babel}

\colorlet{mylightblue}{blue!5!white}
\colorlet{mydarkblue}{blue!30!black}
\colorlet{myblue}{blue!50!black}
\colorlet{myred}{red!50!black}
\colorlet{mydarkred}{red!30!black}
\colorlet{mydarkgreen}{green!30!black}

\newcommand{\sh}{\kern-0.08em$^\textbf{\#}$\hspace{-3pt}}
\renewcommand{\b}{\kern-0.06em$\flat$}

\def\BibTeX{{\rm B\kern-.05em{\sc i\kern-.025em b}\kern-.08em
    T\kern-.1667em\lower.7ex\hbox{E}\kern-.125emX}}

\usepackage{longtable}
\usepackage{setspace}
\usepackage{supertabular}

\usepackage{flushend}
\usepackage{soul}

\usepackage{multirow,hhline,graphicx,array}
\newcolumntype{M}[1]{>{\centering\arraybackslash}m{#1}}

\makeatletter
\def\onecole{%
	\global\columnwidth\textwidth
	\global\hsize\columnwidth
	\global\linewidth\columnwidth
	\global\@twocolumnfalse
	\col@number \@ne
	\@floatplacement}
\makeatother

\newcommand{\STAB}[1]{\begin{tabular}{@{}c@{}}#1\end{tabular}}
\newcommand{\NEWSTAB}[1]{\STAB{1}}
\begin{document}
	
\title{Organization of machine learning based product development as per ISO 26262 and ISO/PAS 21448}
\author{\IEEEauthorblockN{Krystian Radlak}
\IEEEauthorblockA{\textit{exida}, Warsaw, Poland \\
\textit{Silesian University of Technology} \\Gliwice, Poland\\
kradlak@exida.com}
\and
\IEEEauthorblockN{Michał Szczepankiewicz}
\IEEEauthorblockA{\textit{exida} \\
Warsaw, Poland \\
msz@exida.com}
\and
\IEEEauthorblockN{Tim Jones}
\IEEEauthorblockA{\textit{exida} \\
Erlangen, Germany \\
tim.jones@exida.com}
\and
\IEEEauthorblockN{Piotr Serwa}
\IEEEauthorblockA{\textit{exida} \\
Warsaw, Poland \\
pserwa@exida.com}
}


\maketitle


\begin{multicols}{2}
	
\begingroup
\twocolumn
\begin{abstract}
Machine learning (ML) algorithms generate a continuous stream of success stories from various domains and enable many novel applications in safety-critical systems. With the advent of autonomous driving, ML algorithms are being used  in the automotive domain, where the applicable functional safety standard is ISO 26262.
However, requirements and recommendations provided by ISO 26262 do not cover specific properties of machine learning algorithms.
Therefore, specific aspects of ML (e.g., dataset requirements, performance evaluation metrics, lack of interpretability) must be addressed within some work products, which collect documentation resulting from one or more associated requirements and recommendations of ISO 26262. In this paper, we propose how key technical aspects and supporting processes related to development of ML-based systems can be organized according to ISO 26262 phases, sub-phases, and work products. We follow the same approach as in the ISO/PAS 21448 standard, which complements ISO 26262, in order to account for edge cases that can lead to hazards not directly caused by system failure.
\end{abstract}


\endgroup


\section{Introduction}

Parallel with the increased use of machine learning (ML) algorithms, product quality has improved in a variety of areas, including natural language processing \cite{YoungHazarika2018}, speech recognition \cite{NassifShahin2019}, handwritten text recognition \cite{IngleFujii2019}, and computer vision \cite{LiuWand2017} over the last decade. The generalization capabilities of those algorithms enable them to learn and work with incomplete knowledge, thereby making them highly desirable for addressing complex problems for which analytical solutions may not be applicable within an acceptable timeframe or may not even be known. This has motivated the introduction of ML techniques into the manufacture of products  from many industry domains, including automotive systems.

\par 
In the automotive industry, ML-based algorithms are viewed as extremely promising in multiple components of autonomous driving (AD), as they allow vehicles to interpret and comprehend their surrounding environments and make driving-related decisions based on this understanding\cite{GrigorescuTrasnea2020}. Since the malfunction of such systems may cause death or injury, they are safety-related, and so deployment of ML algorithms into such applications must follow a rigorous development lifecycle based on state-of-the-art knowledge and applicable safety standards. 
Recently, the UL 4600 \cite{UL4600} standard has addressed the usage of ML algorithms to evaluate autonomous products, but application of ML techniques is not covered by existing automotive safety standards such as ISO 26262 \cite{ISO-26262-2.0} or ISO/PAS 21448 \cite{ISO-21448}. In addition, the parent standard of ISO 26226, IEC 61508 \cite{IEC-61508}, which is designed for electrical/electronic/programmable electronic safety-related systems, explicitly  recommends not using artificial intelligence algorithms. For this reason, existing automotive safety standards and their verification/validation techniques cannot be directly applied to ML algorithms, as these techniques do not properly address the special characteristics of ML-based components, including their lack of design specification, instability, non-transparency, and limited performance \cite{SalayQueiroz2018,KuwajimaYasuoka2020,SchwalbeSchels2020}.

For automotive AD/ML projects classified as safety-related, ISO 26262 must be followed, both at the process (e.g., change management, configuration management) and technical (e.g., how to build and verify safe software or hardware) levels. Thus, ISO 26262 defines a well-known structure for safety-related automotive projects, and so either all ISO 26262 work products are to be provided or a sufficient argument must be advanced as to why a certain ISO 26262-mandated  step or work product is not necessary. Some of them  need to be extended to cover specific aspects of AD/ML (e.g., datasets requirements, special analyses for ML). Moreover, integration of various software components (e.g., standard C-implemented components and ML code) is also required.

At the present moment, academic approaches do not provide a complete solution due to an incomplete understanding of ML algorithms, including deep neural networks \cite{huang2018survey}. Moreover, the typical focus of most academic work is performance issues, neglecting the safety aspects of the work (e.g., potential failure modes and safety measures that could mitigate potential faults of algorithms and data). Additionally, the whole lifecycle of ML algorithm development (e.g., data collection, the labeling process, algorithm design and development, and model evaluation and monitoring) is typically not taken into account. For example, together with KITTI   \cite{Geiger2012CVPR,Fritsch2013ITSC}, one of the most popular dataset for 3D object detection, its authors  provide a benchmark that allows ranking of object detection algorithms according to an average precision metric. Whereas the researchers  that have published their ML-based solutions focused only on this single evaluation metric, some of the current limitations with respect to published algorithms might be due to insufficient quality of object labeling, e.g., missing object annotations, improper distribution of training and testing datasets, or insufficiencies in parameter optimization. Complete analysis of the whole lifecycle of a concrete ML component from item-level hazard analysis to deployment of binary code on a hardware platform typically exceeds the resources of an academic research group. For instance, in \cite{ZendelMurschitz2015}, the authors have, to date, identified 1470 individual hazards that could affect computer vision algorithms. Coverage of these hazards by the test dataset, definition of safety mechanisms to mitigate these hazards, and integration of these mechanisms into the developed algorithm are required in order to effectively eliminate long-term risks or reduce them to an acceptable level. 

The main goal of this paper is to propose a means of orienting the development process of the very complex ML-based components used in a safety-related context towards well-known ISO 26262 practices that have been utilized for years, instead of creating a completely new standard not yet exposed to the automotive industry. In this paper, we present how existing work products defined in ISO 26262 can cover the additional activities and problems originating from properties of ML algorithms, beginning with item definition and ending with safety validation. Additionally, to illustrate the usage of the proposed methodology, we provide a simplified example that presents the integration of development of an ML-based road lane detection system into existing work products.

In short, by addressing technical key aspects related to the development of ML-based systems, this document provides guidance. Based on evidence compiled from work products and activities performed during development according to ISO 26262 and SOTIF, the proposed approach may prove beneficial for product suppliers and add further assurance that functional safety is achieved for items or elements. However, detailed and complete guidance on all aspects of a safety lifecycle such as performance and pass/fail safety criteria is out of scope of the research described in this article.




The paper is structured as follows: Section II describes existing automotive standards and research including safety assurance and use of ML in the automotive industry. Section III presents our proposals on the means and rationale for integrating the development of ML components used in a safety-related context in AD into existing work products defined in ISO 26262. Also included here is a detailed mapping identifying the work products that require special attention when a vehicle is extended with a new item that uses ML algorithms. Finally, Section IV provides conclusions.



\section{Related work and existing standards}

\par{The automotive industry presently uses ISO 26262 \cite{ISO-26262-2.0} for functional safety. Car manufacturers expect suppliers to use processes designed and operated in accordance with ISO 26262 and to provide components that meet this standard’s requirements. However, it does not address how to define the safety design of autonomous driving vehicles. 
	
The recently published ISO/PAS 21448 standard \cite{ISO-21448} addresses safety of the intended functionality (SOTIF) for automation levels 1 and 2 as defined in SAE J3016 \cite{SAEJ3016}.} The main purpose of ISO/PAS 21448 is to cover foreseeable  misuse of a system by a driver, as well as product system/technological shortcomings. The ISO 26262 standard addresses malfunctions of E/E components, and security standard ISO/SAE 21434 covers the risk associated with a deliberate impact \cite{ISO-21434}. Organizing the development process as per ISO/PAS 21448, ISO 26262, and ISO/SAE 21434 and utilizing their globally well-established and state-of-the-art practices are key to achieving a safe/secure/dependable autonomous system. However, these standards do not present solutions on how to ensure safety  of systems extended by ML-based components.

One of the first attempt to develop a new standard was made by Koopman et al. in \cite{KoopmanFerrell2019}, who proposed  an initial draft of a UL 4600 safety standard for the evaluation of autonomous products to cover autonomous driving and, eventually, other related domains. The UL 4600 standard was released in April, 2020 \cite{UL4600}, and covers the safety principles, tools, techniques, and lifecycle processes employed to design and develop fully automated products. UL 4600 defines the topics that must be addressed in creating and assessing the safety case. It recommends no specific technology to be used and no performance or fail/pass criteria that must be met to produce a safe product. Moreover, UL 4600 emphasizes conformance to the ISO 26262 standard but currently presents only a partial mapping to some clauses from ISO 26262 Part 8, and from ISO 26262 Part 5, related to change management and to hardware development, respectively.

\par A white paper entitled {\it Safety First for Automated Driving} (SaFAD) provides another interesting overview of problems associated with application of ML methods in autonomous driving \cite{safad2019}.  Produced by 11 automotive companies and key technology providers, this technical report summarizes the development and validation of a safe automated driving system. It divides the development of ML components into four steps: (1) define, (2) specify, (3) develop \& evaluate, and (4) deploy \& monitor; provides some technical details on specific solutions or mechanisms; and highlights challenges that need to be addressed. However, it does not map to ISO 26262 or define a precise work product set.

\par Recently, the VDE AR E 2842-61-2 \cite{VDE2842} standard was introduced as a generic framework for the development of trustworthy autonomous systems. The product development lifecycle it defines is similar to that of the IEC 61508 functional safety standard \cite{IEC-61508}, including development descriptions on the solution, system, and technology levels. It further describes splitting management activities related to the development of autonomous systems into three parts: management at the company level, management during the project, and management following release. However, this standard does not define acceptance levels for residual risks and target failure rates, does not  provide ethical guidelines, does not introduce specific requirements related to any trustworthiness aspect of the application or technology, and, most importantly, does not employ the ISO 26262 standard’s work product organization.
\par Also worth mentioning is that ISO has created a new technical subcommittee (ISO/JTC 1 SC 42 \cite{ISOSC42}) to operate in the area of artificial intelligence. Specifically, the scope of this subcommittee’s work  covers fundamental standards as well as issues related to safety and trustworthiness \cite{Rudzicz2019}. Another  standard under development is ISO/CD TR 4804 \cite{ISO-4804}, whose aim is to cover safety and cybersecurity aspects for automated driving systems.

\par The topic of safety assurance of ML algorithms and their usage in autonomous vehicles continues to gain increased attention from the scientific community, and several interesting papers in this area have been published. The papers in the first such group attempt to analyze the possibility of adaption and extension of existing functional safety standards such as ISO 26262. Salay et al. in \cite{SalayQueiroz2018, SalayCzarnecki2018} presented the first attempt at an impact analysis of ML-algorithm usage on various parts of ISO 26262. In these works, the authors discussed five topics that should be addressed in development of ML-based components such as the following: new types of hazards specific to ML, new types of faults and failure modes, usage of incomplete training datasets, the level on which ML algorithms should be used, and which software techniques should be required in these algorithms verification. They also analyzed the possibility of using unit-level software techniques that ISO 26262 highly recommends and concluded that 70\% of them can by directly applied or adopted for development of ML-based components. However, their analysis does not cover the effectiveness level of these software techniques in covering newly identified failure modes and hazards.
\par In \cite{HenrikssonBorg2018}, Henriksson et al. propose that the ML component can be realized as a software unit on the software level phase and discuss the most critical gaps between ISO 26262 and ML development. Additionally, they propose three adaptations related to ML training, model sensitivity, and test case design that are crucial to development of ML-based components according to ISO 26262. We do not follow this approach, as the development of ML needs to start at item level.
\par Recently, Schwalbe and Schels in \cite{SchwalbeSchels2020} proposed organizing ML development as a top-down approach. They also presented a summary of existing methods supporting the safety argumentation of an ML-based system and discuss major open challenges such as the following: requirement engineering for ML, prevention of faults through proper design of ML models based on expert knowledge, verification of the ML model against defined requirements and test data, and finally validation of requirement and ML test case completeness. However, the proposed approach  is only aligned with the ISO 26262 Part 6, process and also does not provide a direct mapping to work products.

\par The second group of papers attempt  to define the open challenges to be addressed in the development of ML-based components used in autonomous driving from a functional safety perspective. Additionally, some propose solutions to cover existing gaps but fail to provide a complete solution to mitigate all risks introduced by ML algorithms.





\par Kuwajima et al. \cite{KuwajimaYasuoka2020} analyzed open challenges in engineering safety-critical systems that employ ML methods. These authors identified the following factors as having the greatest impact on quality of ML models: lack of requirement specifications (i.e., specification of preconditions for the system itself and of detail level for functions) and lack of robustness (due primarily to uncertainty of observation and extremely low probability of edge cases). Other identified limitations of ML systems in comparison to conventional systems are  lack of design specification and lack of interpretability, although these are classified as less important.

\par Spanfelner et al. \cite{Spanfelner2012} highlighted that perception functionality for autonomous driving may not be completely specifiable and that human categories (e.g., pedestrians) can only be partially specified using rules (e.g., necessary and sufficient conditions) and also require examples. The authors in \cite{SalayCzrnecki2019} also addressed this topic, proposing the use of partial specifications instead of complete ones, whose precise specification is highly difficult and, in some cases, impossible.
\par  Mohseni et al. in \cite{MohseniPitale2020} reviewed and categorized several techniques that can be used to enhance dependability and safety of ML algorithms. In the scope of this work, they analyzed different error detection mechanisms such as uncertainty estimation methods, in-distribution and out-of-distribution error detectors, and techniques to measure algorithm robustness.
\par  Willers et al. in \cite{WillersSudholt2020} defined safety concerns and associated mitigation approaches regarding deep learning methods used in autonomous agents. These concerns are related to the following issues: failure of data distributions to adequately approximate real-world distributions; distributional shifts in data over time; incomprehensible behavior; unknown behavior in rare, critical situations; unreliable confidence information; brittleness of deep neural networks (DNNs); inadequate separation of test and training data; dependence on labeling quality; and insufficient consideration of safety in metrics.
\par Cheng et al. in \cite{ChengHuang2018} introduced a novel tool called the NN-Dependability Kit to support safe design and development of DNNs for use in autonomous driving systems. This tool offers several state-of-the-art techniques that could improve safety engineering of DNNs, including dependability metrics, techniques for ensuring that the generalization does not lead to undesired behaviors, and runtime monitoring methods.

\section{Proposed organization of ML-based product development}

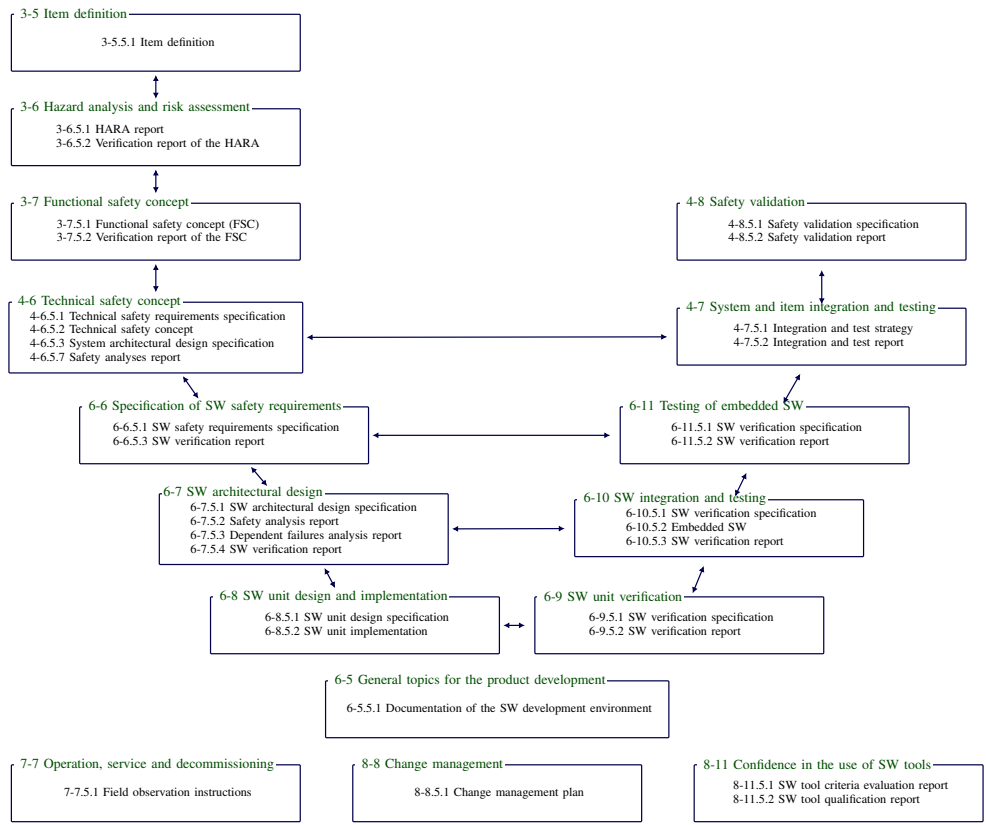
\begin{figure*}\centering
\begin{tikzpicture}[xscale=1.3,yscale=1.2,
                    fscale/.style={font=\relsize{#1}},
                    arrow/.style={->,thick,mydarkblue,shorten <=2,shorten >=4},
                    sarrow/.style={->,thick,mydarkblue,shorten <=2,shorten >=6}]
  \def\c{0.7}
  \def\m{5}
  \def\tuning(#1,#2,#3,#4,#5){
    \node[draw=mydarkblue,thick,rounded corners=\c,inner sep=\m, minimum width=6.6cm, minimum height=1.3cm] (#1) at (#2,#3) [fscale=1.2,anchor=south] { {\footnotesize  #4}};
    \node[fill=white,inner sep=1] at (#1.north west) [fscale=-2,right=2,anchor=west,text=mydarkgreen] {{\small #5}};
  }
 
  \tuning(A,  1, 5.2, \begin{tabular}{l}  3-5.5.1 Item definition 
      \end{tabular},     3-5 Item definition)
  
  \tuning(B,  1, 3.4,  \begin{tabular}{l}   3-6.5.1 HARA report \\
 3-6.5.2 Verification report of the HARA
      \end{tabular},      3-6 Hazard analysis and risk assessment )
 
 \tuning(C,  1, 1.6, \begin{tabular}{l} 3-7.5.1 Functional safety concept (FSC)\\
  3-7.5.2 Verification report of the FSC
      \end{tabular},      3-7 Functional safety concept)
  
  \tuning(D,  1, -0.55, \begin{tabular}{l}  4-6.5.1 Technical safety requirements specification \\
  4-6.5.2 Technical safety concept \\
  4-6.5.3 System architectural design specification \\
  4-6.5.7 Safety analyses report
      \end{tabular},      4-6 Technical safety concept)
  
    \tuning(E,  7, -7.5,  \begin{tabular}{l}  6-5.5.1 Documentation of the SW development environment
      \end{tabular} ,      6-5 General topics for the product development)
    
      \tuning(F,  2.2, -2.28, \begin{tabular}{l}  6-6.5.1 SW safety requirements specification \\
      6-6.5.3 SW verification report
      \end{tabular} ,      6-6 Specification of SW safety requirements)
      
      \tuning(G,  3.6, -4.2,\begin{tabular}{l}  6-7.5.1 SW architectural design specification \\
      6-7.5.2 Safety analysis report \\
      6-7.5.3 Dependent failures analysis report\\
      6-7.5.4 SW verification report   
      \end{tabular},6-7 SW architectural design)
      
      \tuning(H,  4.5, -5.9,  \begin{tabular}{l}  6-8.5.1 SW unit design specification \\
      6-8.5.2 SW unit implementation \\
      \end{tabular},      6-8 SW unit design and implementation)
      
       \tuning(I,  10.2, -5.9, \begin{tabular}{l}  6-9.5.1 SW verification specification \\
       6-9.5.2 SW verification report \\
      \end{tabular} ,       6-9 SW unit verification)
       
       \tuning(J,  10.9, -4.05, \begin{tabular}{l}  6-10.5.1 SW verification specification \\ 
       6-10.5.2 Embedded SW \\
       6-10.5.3 SW verification report
      \end{tabular} ,       6-10 SW integration and testing)
       
       \tuning(K,  11.7, -2.28, \begin{tabular}{l}  6-11.5.1 SW verification specification  \\
       6-11.5.2 SW verification report
      \end{tabular} ,      6-11 Testing of embedded SW)
       
        \tuning(L,  12.7, -0.4, \begin{tabular}{l}  4-7.5.1 Integration and test strategy \\
        4-7.5.2 Integration and test report
      \end{tabular} ,      4-7 System and item integration and testing)
        
         \tuning(M,  12.7, 1.6, \begin{tabular}{l}  4-8.5.1 Safety validation specification  \\
         4-8.5.2 Safety validation report \end{tabular} ,      4-8 Safety validation)
       
       	\tuning(E,  1.0, -9.1, \begin{tabular}{l}  
       7-7.5.1 Field observation instructions
       \end{tabular},      7-7 Operation, service and decommissioning)
       
       \tuning(E,  7.0, -9.1, \begin{tabular}{l}  
       8-8.5.1 Change management plan
       \end{tabular},      8-8 Change management)
       
              \tuning(E,  13.0, -9.1, \begin{tabular}{l}  
			8-11.5.1 SW tool criteria evaluation report \\
			8-11.5.2 SW tool qualification report
       \end{tabular},      8-11 Confidence in the use of SW tools)

  \draw[sarrow,<->] (A)  -- (B);
  \draw[sarrow,<->] (B)  -- (C);
  \draw[sarrow,<->] (C)  -- (D);
  \draw[sarrow,<->] (D)  -- (F);
  \draw[sarrow,<->] (F)  -- (G);
  \draw[sarrow,<->] (G)  -- (H);
  \draw[sarrow,<->] (H)  -- (I);
  \draw[sarrow,<->] (I)  -- (J);
  \draw[sarrow,<->] (J)  -- (K);
  \draw[sarrow,<->] (K)  -- (L);
  \draw[sarrow,<->] (L)  -- (M);
  
  \draw[sarrow,<->] (D)  -- (L);
    \draw[sarrow,<->] (F)  -- (K);
 \draw[sarrow,<->] (G)  -- (J);
 
\end{tikzpicture}
 
	\caption{ISO 26262 sub-phases and work products affected by the usage of ML algorithms.} \label{v_model}
\end{figure*}

\begin{figure*}\centering
	 
\setlength{\tabcolsep}{2pt}
\renewcommand{\arraystretch}{0.5}
 
\begin{tikzpicture}[xscale=1.3,yscale=1.2,
                    fscale/.style={font=\relsize{#1}},
                    arrow/.style={->,thick,mydarkblue,shorten <=2,shorten >=4},
                    sarrow/.style={->,thick,mydarkblue,shorten <=2,shorten >=6}]
  \def\c{0.7}
  \def\m{5}
  
  \def\tuning(#1,#2,#3,#4,#5){
    \node[draw=mydarkblue,thick,rounded corners=\c,inner sep=\m, minimum width=6cm, minimum height=1.4cm] (#1) at (#2,#3) [fscale=1.2,anchor=south] { {\footnotesize  #4}};
    \node[fill=white,inner sep=1] at (#1.north west) [fscale=-2,right=2,anchor=west,text=mydarkgreen] {{\small #5}};
  }
 \def\newtuning(#1,#2,#3,#4,#5){
	\node[draw=mydarkblue,thick,rounded corners=\c,inner sep=\m, minimum width=8.5cm, minimum height=1.4cm] (#1) at (#2,#3) [fscale=1.2,anchor=south] { {\footnotesize  #4}};
	\node[fill=white,inner sep=1] at (#1.north west) [fscale=-2,right=2,anchor=west,text=mydarkgreen] {{\small #5}};
}

  \tuning(A,  1.4, 5.3, \begin{tabular}{l}  SOTIF-based system properties for the item 
      \end{tabular},     3-5 Item definition)
  
  \tuning(B,  1.4, 3.45,  \begin{tabular}{l}  HARA for ML \\
  Safety goals for ML 
      \end{tabular},      3-6 Hazard analysis and risk)
 
 \tuning(C,  1.4, 1.6, \begin{tabular}{l}  Functional safety requirements  for ML 
      \end{tabular},      3-7 Functional safety concept)
  
  \tuning(D,  1.4, -0.25, \begin{tabular}{l} Technical safety requirements for ML  \\
  Technical safety concept for ML \\
  System architectural for ML \\
  Safety analyses report for ML 
      \end{tabular},      4-6 Technical safety concept)

      \tuning(F,  2.5, -2, \begin{tabular}{ll} {\bf DATASET:} &  dataset and labelling policy specification and review  \\  [2pt]  \hline \\[-1pt] 
      {\bf  MODEL:} & requirements and review for ML algorithm(s){,} \\ 
      & KPI and runtime monitoring specification and review
      \end{tabular} ,      6-6 Specification of software safety requirements)
      
      \tuning(G,  3.2, -3.95,\begin{tabular}{ll}  
      {\bf DATASET:} & architectural design of dataset and labelling policy\\ [2pt]  \hline \\[-1pt] 
      {\bf MODEL:} & architectural design of ML algorithm(s) for host and target\\
       & architectural design of KPI  and runtime monitoring \\
      &  integration with rest of software
      \end{tabular},       6-7 Software architectural design)
      
      \newtuning(H,  3.9, -6.15,  \begin{tabular}{ll}   
       {\bf DATASET:}  & dataset gathering and labelling \\ [2pt]  \hline \\[-1pt] 
      {\bf MODEL:} & 
        implementation of host and target models \\
	  & implementation of host and target frameworks \\
      & implementation of KPI and runtime monitoring\\
      & training and validation{,} parameters optimization   
      \end{tabular},      6-8 Software unit design and implementation)
      
       \tuning(I,  11.9, -6.15, \begin{tabular}{ll}  
       {\bf DATASET:} & dataset labelling quality of the verification specification and report\\ [2pt]  \hline \\[-1pt] 
      {\bf MODEL:}  & training and validation verification specification and report \\
      & back-to-back testing between target model and host model \\
      & testing performance of the elementary models using KPI \\
      & testing runtime monitoring
       \end{tabular} ,       6-9 Software unit verification)
       
       \tuning(J,  12.6, -3.95, \begin{tabular}{ll}  
       {\bf DATASET:} &  verification of the consistency between subsets of annotations\\ [2pt]  \hline \\[-1pt] 
      {\bf MODEL:} & integration of ML  into bigger ones \\
       & testing of integrated ML models using KPI\\ 
       & testing models together with runtime monitoring \\
      \end{tabular} ,       6-10 Software integration and testing)
       
       \tuning(K,  13.2, -2, \begin{tabular}{ll}  
       {\bf DATASET:} &  dataset completeness{,} coverage and equivalence classes\\ [2pt]  \hline \\[-1pt] 
      {\bf MODEL:}  & E2E-testing of entire software \\
       & testing if ML component fulfils the SSRs
      \end{tabular} ,      6-11 Testing of embedded software)
       
        \tuning(L,  14.4, -0.25, \begin{tabular}{l}  HW-SW testing/integration on target vehicle
      \end{tabular} ,      4-7 System and item integration and testing)
        
         \tuning(M,  14.4, 1.6, \begin{tabular}{l} Specification of the road test \\
         Execution of road tests\end{tabular} ,      4-8 Safety validation)

    \tuning(E,  8, -7.8, \begin{tabular}{l}  
      Selection of the ML development environment (ML frameworks and tools for simulation and testing)
    \end{tabular},      6-5 General topics for the product development)
 
%
 	
       	\tuning(E,  2.0, -9.5, \begin{tabular}{l}  
Continuous monitoring of anomalies in ML-based system
\end{tabular},      7-7 Operation, service and decommissioning)

\tuning(E,  7.6, -9.5, \begin{tabular}{l}  
Change management for continous delivery
\end{tabular},      8-8 Change management)

\tuning(E,  13.6, -9.5, \begin{tabular}{l}  
Evaluation of TCL for ML lools
Qualification of specific ML tools
\end{tabular},      8-11 Confidence in the use of software tools)

  \draw[sarrow,<->] (A)  -- (B);
  \draw[sarrow,<->] (B)  -- (C);
  \draw[sarrow,<->] (C)  -- (D);
  \draw[sarrow,<->] (D)  -- (F);
  \draw[sarrow,<->] (F)  -- (G);
  \draw[sarrow,<->] (G)  -- (H);
  \draw[sarrow,<->] (H)  -- (I);
  \draw[sarrow,<->] (I)  -- (J);
  \draw[sarrow,<->] (J)  -- (K);
  \draw[sarrow,<->] (K)  -- (L);
  \draw[sarrow,<->] (L)  -- (M);
  
  \draw[sarrow,<->] (C)  -- (M);
  \draw[sarrow,<->] (D)  -- (L);
  \draw[sarrow,<->] (F)  -- (K);
  \draw[sarrow,<->] (G)  -- (J);

\end{tikzpicture}
 
	\caption{ISO 26262 sub-phases and proposed aspects of ML that need to be addressed during development of  ML-based components. \label{ml_v_model}}
\end{figure*}
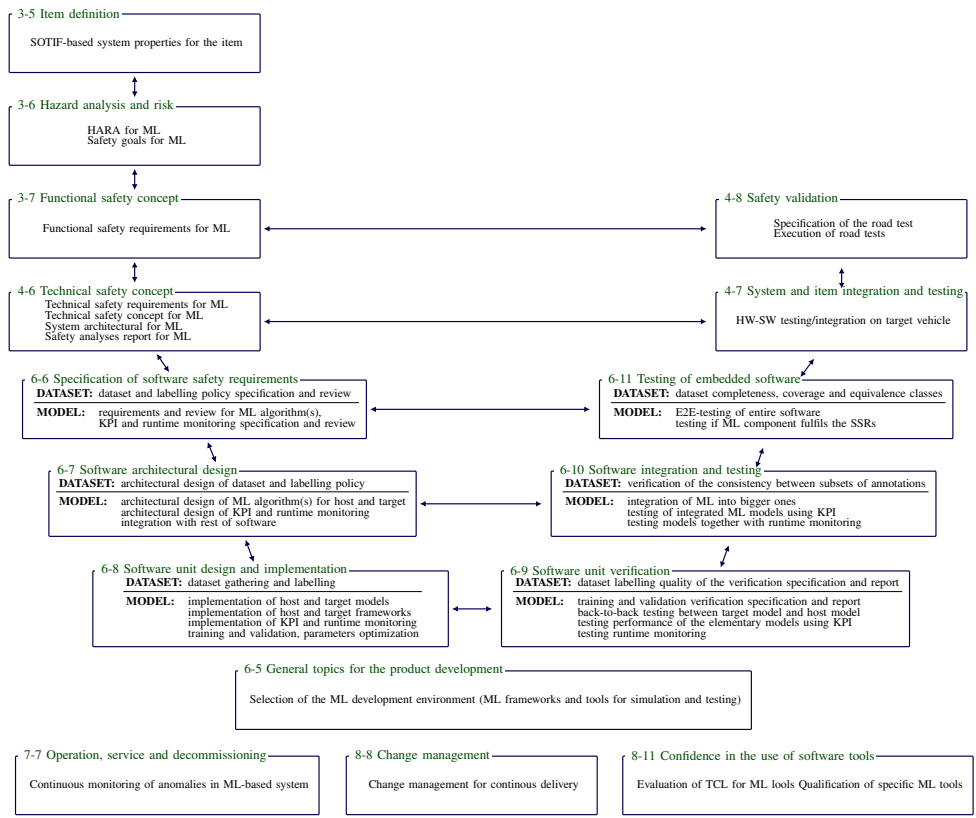

Providing a dedicated standard that would address all functional safety aspects of ML-based components used in a safety-related context within the automotive domain would entail expenditures of large amounts of work and time on its creation, the required careful review process, and the negotiations needed to reach an agreement across the entire safety-expert community. It could also introduce redundancies or contradictions in standards already established. Therefore, instead of delivering a stand-alone standard, the comprehensive solution we propose is aligned with the well-known and widely applied ISO 26262. We recommend this approach for to following reasons:
\begin{itemize}
    \item ISO 26262 is a well-established standard with well-defined terminology and has been used by automotive industry and functional safety engineers since 2011.
    \item Our proposal would easily integrate with and fit into the existing processes and procedures defined by ISO 26262.
    \item Since components that use ML algorithms are specific software modules that interface with other safety-related software components and run on safety-related hardware, they have to be designed, integrated, validated, and tested with other components in accordance with ISO 26262.
    \item It unifies the methodology of development for both ML-based and classical components used in vehicle functions.
    \item ISO 26262 does not provide detailed information as to how the requirements for concrete functionalities are to be defined, e.g., how to provide safe, persistent storage or safe CPUs, and so complements our approach, which does not provide detailed guidance on ML development.
    \item A detailed guidance and good practices recommended to ML can be defined in separate documents as ISO 26262 does not provide any methodology how to achieve the required rigor level. 
    \item ISO 26262 leaves much room for interpretation, especially with regard to the technology to be used, or it refers to test methods without defining them in detail, e.g., the fault injection test. Thus, it would be possible to define equivalent methods for handling ML-specific software.
\end{itemize}


The product-development lifecycle introduced in the ISO 26262 standard is divided into phases and sub-phases for which the required work products are defined. This section points out which sub-phases and work products are affected when ML-based components are to be used in a vehicle and be incorporated into the V-model proposed in ISO 26262. 


\par In the scope of this work, we focus on supervised learning and deep neural networks, as these achieve superior performance in many of the computer vision tasks involved in autonomous driving \cite{SzeChen2017,BadueGuidolini2019,SenguptaBasak2019}. This document therefore does not address problems related to the development of end-to-end learning \cite{bojarski2016end} components, in which DNNs are trained using raw sensors data to directly perform steering of the vehicle.

In contrast to existing works, which recommend organizing the lifecycle of an ML-based product according to the V-model \cite{SalayCzarnecki2018, HenrikssonBorg2018, SchwalbeSchels2020}, we do not limit ML development to the algorithmic solutions and ISO 26262 Part 6. In comparison to our approach, Salay et al. in \cite{SalayCzarnecki2018} assumed that ML is only used to implement individual software units and not entire system.

\par In our work, we propose to include design and development of ML solutions beginning with the conceptual work defined in the {\it ISO 26262 Part 3: Concept Phase}. In this way, a complete solution and its environment are defined using a top-down approach instead of focusing on specific technological aspects of ML algorithms. Then, the indented behavior of the system linking the abstract level of the ML-based product with specific technological solutions and its validation is organized as per {\it ISO 26262 Part 4: Product development at system level}. The algorithmic aspects of ML solutions and their interfaces with other product parts are defined according to {\it ISO 26262 Part 6: Product development at software level}. Additionally, we recommend to adopt other parts of ISO 26262 when ML-based components are used. The affected parts are then the following: the continuous monitoring of a released product after the start of production that {\it ISO 26262 Part 7: Operation, services and decommissioning} specifies and the supporting processes (e.g., change plan management and qualification of tools) explained in {\it ISO 26262 Part 8: Supporting processes}. All existing work products that are affected by introduction of ML-based components into the road vehicle are presented in Fig. \ref{v_model}. The specific proposed additional activities that need to be added to the existing sub-phases defined in ISO 26262 during ML-based component development are summarized in the V-model presented in Fig \ref{ml_v_model}. {\it ISO 26262 Part 5: Product development at the hardware level} is outside the scope of this article; however if an ML algorithm is deployed on a specific (custom) piece of hardware, then the hardware architectural design can be essential for potential optimizations of the algorithm. In such instances, Part 5 of ISO 26262 needs to be taken into account as well.
	
\par In Tab. \ref{tab:myfirstlongtable}, we also introduce a detailed mapping of specific ML-based activities to existing ISO 26262 work products. Additionally, for each work product, we provide an example inspired by a road lane detection system so as to highlight which aspects of ML-based component development should be addressed in which work products. These examples are provided to demonstrate the feasibility of the proposed mapping; however, completeness and analysis-based argumentation of the given examples is beyond the scope of this work. Moreover, verification activities shown on the left side of the V-model depicted in  Fig. \ref{v_model} (i.e., Hazard Analysis \& Risk Assessment (HARA), Functional Safety Concept, SW verification report) are assumed to be able to identify gaps within work products of the corresponding sub-phases, and so those are not explained in detail. 

\par The main benefit of our proposal is that initiating ML activities in the concept phase enables  the expert knowledge of ML engineers to be involved from the beginning. Such an approach avoids limiting potential solutions by purely technological and design aspects established at later phases of the development, e.g., selection of incapable sensors or limited hardware. It also facilitates addressing cases in which architects on the system level propose a technical safety concept, which do not provide specific hardware resources required to develop specific ML solutions, e.g., two diverse DNNs would require two independent components with GPU or hardware accelerators. In cases where ML activities are started solely at the software level of ISO 26262 Part 6, insufficient hardware resources could block modification of specific algorithmic solutions.

\par Defining the whole process according to the workflow proposed by ISO 26262 makes possible easily tracing potential failures and limitations in product development. Thus, it is very beneficial and can facilitate identification of fundamental complex flaws. For instance, in the Automated Lane Keeping System, the proposed approach would help to identify incompleteness of dataset requirements during validation at system level, even if model performance on the software level was evaluated to be sufficient.
\subsection{Limitation of the proposed work}

The main limitation of this work is the complexity of the validation of the proposed approach. In our proposal, ML system is developed starting from writing higher-level to lower-level  requirements instead of using a fixed dataset. In order to perform a full validation of the proposed approach, it is necessary to develop ML algorithm together with a collection of the dataset basing on specified requirements. Finally, the developed system is validated during road tests, whereas the dataset and ML algorithm is enhanced using samples acquired during tests. Therefore, the proposed process-related solution for the development of ML algorithms for safety-critical systems is a long-term contribution. Coverage of all related aspects is beyond even the capabilities of existing manufacturers and would require consolidated cooperation between the automotive industry and academia. Our approach also needs to be consulted and accepted by ML engineers and functional safety authorities. Therefore this work may be considered as a preliminary proposal on how ML-based components can be developed using existing ISO 26262 and ISO/PAS 21448 before a dedicated standard is developed.

\end{multicols}

\begingroup
\onecole
\begingroup
\smaller

\newcommand{\linewrapcol}[1]{\parbox[c][][l]{0.16\textwidth}{\begin{flushleft}#1\end{flushleft}}}

\begin{longtable}{|||p{0.16\textwidth}|m{0.78 \textwidth}|||} 
	
	\caption{Proposed allocation of ML activities as per ISO26262 work products. The cells with exemplary activity for each work product were highlighted using gray color.} \label{tab:myfirstlongtable}\\
	\endfirsthead
	\caption* {Table \ref{tab:myfirstlongtable} (continued)} 
	\endhead
	\endfoot

	\hline
     Work product & Interpretation of ISO 26262 / ISO/PAS 21448 work products for ML \\ 
    \hline \hline
    
   	\multicolumn{2}{|c|}{Part 3 Concept phase} \\ \cline{1-2}
   \multicolumn{2}{||c||}{Clause 3-5 Item definition} \\ \cline{1-2}

       \multirow{2}{*}{\linewrapcol{3-5.5.1 Item definition}} & Define the new interfaces, boundaries, item functions, and item properties that appear due to ML being employed at the vehicle level and the operational design domain (ODD) in which the autonomous driving functionality is designed to properly operate (i.e., environmental conditions, background scene, geographic domain, speed limit, etc.). It also needs to address the SAE J3016 standard by identifying the automation level. \\ \hhline{|||~-}
       & \cellcolor{gray!25}ITEM: SAE L3 Automated Lane Keeping System. \newline 
       VEHICLE FUNCTION: vehicle shall keep the lane without lane changing under defined ODD conditions.
       \newline TAILORING: include the computational infrastructure that performs the online/incremental ML update function.
       \newline DESCRIPTION OF ITEM: Data from radars, lidars, and cameras shall be included in the decision-making process related to controlling engine, brakes, and steering.
       \newline ODD: driving on a dry highway without rain or snow during daylight with 120 km/h max. speed limit. 
       \\ \cline{1-2}

	\multicolumn{2}{||c||}{Clause 3-6  Hazard analysis and risk assessment} \\ \cline{1-2}     
   		\multirow{2}{*}{\linewrapcol{3-6.5.1 Hazard analysis and risk assessment report}} &  
		Identify hazards  caused by the unintended behaviour of the functionality which uses ML algorithms to cover functional limitations of ML and  formulate related safety goals. \\ \hhline{|||~-}
   	   & \cellcolor{gray!25}HAZARD H01: highway pilot incorrectly detects road lane due to model limitations, e.g. new pattern.
        \newline SAFETY GOAL SG01: Prevent lane departure.
        \\ \cline{1-2}
         
        
    	\multicolumn{2}{||c||}{Clause 3-7  Functional safety concept} \\ \cline{1-2}  
	     \multirow{2}{*}{\linewrapcol{3-7.5.1 Functional safety concept}}  & 
		    Define high-level safety requirements that are specific to ML in order to ensure avoidance or control of relevant faults in accordance with their  safety goals and assign them to the respective elements in the system architectural design.  \\ \hhline{|||~-}
    	&  \cellcolor{gray!25}FUNCTIONAL SAFETY REQUIREMENTS: 
    	\newline FSR 01: The vehicle shall recognize new road scenes using runtime monitoring system.
    	\newline FSR 02: The detected new scenes shall be sent to safe server farms.
    	\newline FSR 03: The current road lane detection model shall be updated using the new annotated scenes.
    	\newline FSR 04: The updated model shall be sent to the target vehicle.
        \newline ALLOCATION TO ELEMENT: Allocated to the perception module and primary processing channel.
	    \\ \cline{1-2}
    
        
        \multicolumn{2}{|c|}{Part 4 Product development at the system level} \\ \cline{1-2}
        
   		\multicolumn{2}{||c||}{Clause 4-6 Technical safety concept} \\ \cline{1-2} 
      \multirow{2}{*}{\linewrapcol{4-6.5.1 Technical  safety requirements specification}} & Define technical safety-related requirements specific to ML usage. \\ \hhline{|||~-} & \cellcolor{gray!25}TSR01: Camera ECU shall receive raw data from camera sensor at the rate of 30 frames per second.
	    \newline TSR02: Camera ECU shall convert, filter, process input data and provide it every 10ms over Ethernet to the primary processing channel.   \\ \cline{1-2}
  
	    \multirow{2}{*}{\linewrapcol{4-6.5.2 Technical safety concept}} & Define technical safety concept that specifies how system architectural design fulfills safety requirements, given known limitations of ML. \\ \hhline{|||~-} & \cellcolor{gray!25}Road lane detection shall be performed using  two diverse ML algorithms that employ separate cameras. The first method shall work on the primary processing channel that utilizes GPU and the latter on the secondary processing channel. Voting shall be applied to decide between the results of these algorithms. \\ \cline{1-2}
    
	    \multirow{2}{*}{\linewrapcol{4-6.5.3 System architectural design specification}} & Define system architectural design that shall realize the technical safety requirements, given known limitations of ML algorithms. \\ \hhline{|||~-} & \cellcolor{gray!25}Specification of the component architecture (primary, secondary, cameras), specification of their interfaces and format of exchanged data, specification of signal/information flow, specification of system-level state machines (e.g., autonomous mode enabled or disabled), and system-level partitioning of the ML functionality into sense-plan-act system components \cite{safad2019}. \\ \cline{1-2}
    
  	  	\multirow{2}{*}{\linewrapcol{4-6.5.7 Safety analyses report }} & Provide a safety analysis on the system level that verifies each system-level ML component and that demonstrates the independence between independent elements. \\ \hhline{|||~-} &  \cellcolor{gray!25}Analysis demonstrated  that both ML components use similar architecture of DNN and were trained using the same training dataset, thus causing level of diversity to be insufficient and prone to erroneous output for the same samples. In this case, it is recommended to replace one DNN architecture with another or with a different ML algorithm having adequate capabilities. \\ \cline{1-2} \pagebreak

	    \cline{1-2} 
    \multicolumn{2}{||c||}{Clause 4-7  System and item integration and testing} \\ \cline{1-2} 
   	 	\multirow{2}{*}{\linewrapcol{4-7.5.1 Integration and test strategy}}   &  Define a methodology for testing on the target vehicle  against technical safety concept given known limitations of components that used ML algorithms. \\ \hhline{|||~-} &   \cellcolor{gray!25}Extensive testing by injecting faults to the input of the primary processing channel to verify whether the fault is detected by the voter defined in the technical safety concept. Extensive testing by providing generated and real images to the primary channel to test  system robustness against artificially generated adversarial attacks.  \\ \cline{1-2}
    
	    \multirow{2}{*}{\linewrapcol{4-7.5.2 Integration and test report}} &  Provide test specifications and a report that covers 
	    integration of individual ML-based elements of the system  on the  target vehicle according to the defined specification. \\ \hhline{|||~-} & \cellcolor{gray!25}During extensive tests, the primary processing channel was found to incorrectly detect some road lane instances. \\ \cline{1-2}  

    \multicolumn{2}{||c||}{Clause 4-8 Safety validation} \\ \cline{1-2} 
	    \multirow{2}{*}{\linewrapcol{\vspace{-0.3cm}4-8.5.1 Safety validation specification including safety validation environment description}} &  Specify ML/AD-related road tests addressing and providing argumentation concerning their coverage and completeness in the context of known ML limitations.
	    \newline
	    \\ \hhline{|||~-} &  \cellcolor{gray!25}Autonomous vehicle shall be tested on the highway under the defined ODD. \newline \\ \cline{1-2}
  
	 \multirow{2}{*}{\linewrapcol{4-8.5.2 Safety validation report resulting from requirements}} & Perform the specified road tests, report results, and make decision whether current version of the product ensures expected level of safety. \\ \hhline{|||~-}
	 & \cellcolor{gray!25}Current version of the highway pilot does not recognize rainy weather conditions and incorrectly allows enabling the autopilot feature outside defined ODD. \\ \cline{1-2}
	\multicolumn{2}{|c|}{Part 6 Product development at the software level} \\ \cline{1-2}
     \multicolumn{2}{||c||}{Clause 6-5 General topics for product development} \\ \cline{1-2} 
    
    	\multirow{2}{*}{\linewrapcol{6-5.5.1 Documentation of the SW development environment}} & Select the ML development environment and its configuration, e.g., frameworks for model training, execution, ML lifecycle management, and data annotation management tools. \\ \hhline{|||~-} & \cellcolor{gray!25}Selected DNN for road lane detection shall be implemented in TensorFlow framework. The training of the model shall use Python, and the runtime model shall be developed in C++. ML lifecycle management shall be handled by MLflow. 
	    \\ \cline{1-2}
    
    \multicolumn{2}{||c||}{Clause 6-6 Specification of software safety requirements } \\ \cline{1-2} 
	    \multirow{2}{*}{\linewrapcol{6-6.5.1 Software safety requirements specification}} & Define general requirements for both the selected ML algorithm and dataset. This work product should cover the following aspects:
	     \newline - specification of the dataset (amount of data, type of data needed (e.g., object classes, ODD, weather conditions, geographic domain, background scene), division of data between training, validation, and testing), \newline
	     - specification of labeling policy (data annotation, treatment of occluded objects, number of annotators annotating the same data), \newline 
	     - specification of KPIs for dataset (labeling quality evaluation, dataset coverage, dataset distribution), \newline 
	     - specification for the ML algorithm (type of algorithm to be used, specific tasks the algorithm should perform, required computational complexity), \newline 
	     - requirements on KPIs for ML model (metrics to measure model performance, e.g., average precision, measure of model robustness with respect to noise, data augmentation, adversarial attacks, reproducibility of results), \newline 
	     - requirements of runtime monitoring (metrics to be run on the target vehicle to prevent  potential failures, e.g., uncertainty metrics, recognition of out-of-distribution  data).
	
		The scope of this work product will probably be one of the most challenging since current knowledge concerning ML and DNN is still insufficient. However, specific software safety requirements related to ML algorithms and dataset can be included as a part of this work product. In \cite{Zendel2017}, challenges and limitations related to dataset design and development are defined, and \cite{feng2019deep} reviewed the algorithms related to perception in autonomous driving. \\ \hhline{|||~-} & \cellcolor{gray!25}DATASET:  The dataset shall contain images acquired for different road types during differing weather conditions.  The road lane boundaries shall be marked pixel by pixel.  Each image shall be annotated by two independent annotators.  The amount of 10\% of randomly selected data shall be additionally annotated by a third annotator.  The data acquisition shall take place during day time.
        \newline MODEL:  Road lane detection shall be performed by an algorithm using two different architectures of deep neural networks.  The lane position deviation  \cite{SatzodaEvaluationMetrics} metric of both DNN architectures shall not exceed the defined threshold. 
        \newline NOTE: The 10\% threshold of the data additionally annotated by the third annotator has been established arbitrarily for the purpose of  the exemplary software safety requirement  related to the dataset. This threshold shall be determined based on calculation of the statistical significance. \\ \cline{1-2}

%
    
    \multicolumn{2}{||c||}{Clause 6-7 Software architectural design } \\ \cline{1-2} 
	    \multirow{2}{*}{\linewrapcol{6-7.5.1 Software architectural design specification}} & Design the software architecture for the components that use ML including preprocessing of input data from sensors, selection of the ML algorithms and the data flow between them, and postprocessing of the outputs of ML algorithms. It shall also define the architectural design of the ML model’s training and its integration with KPI to evaluate the performance and robustness  of the model during training. Additionally, the architecture shall address the problem of integrating the ML model with runtime monitoring metrics, which allows supervision of the ML model’s behavior (e.g., analysis of neuron activation patterns \cite{ChengNuhrenberg2019}) and correctness of input data (e.g., using Monte Carlo epistemic uncertainty to measure statistical similarity of the analyzed sample to samples used in the training dataset \cite{Kendall2017}) at inference time.
	    \newline
	    This part should also cover architectural design of the dataset development process and specify how the data should be recorded, collected, and annotated (data formatting, compression, and labeling). The problem of defining the architectural design for data gathering and the labeling process is typically neglected, but, in our opinion, these steps shall be considered as a part of product development. One possible approach for handling these aspects could be to use active learning \cite{clemensalex2018active, Roy2018} in the development life cycle to label only those data samples which are expected to improve the model’s quality or to tune the model to better deal with edge cases unrecognized during the tests. \\ \hhline{|||~-} 
	    & \cellcolor{gray!25}DATASET: Road lane annotations shall be detected first by pretrained the LaneNet model. Correctness of an automatically detected road lane shall be validated  by an annotator and corrected in case incorrect results from the model were identified.
	    \newline MODEL: The road lane shall be localized using the LaneNet network \cite{wang2018lanenet} and Canny detector \cite{WuChang2014}, and consistency between them should be measured according to the defined KPI to verify correct road lane detection.
	    \\ \cline{1-2} \pagebreak

\cline{1-2} 
     \multirow{2}{*}{\linewrapcol{6-7.5.2 Safety analysis report}} &  Perform safety analysis tailored for ML to confirm that all software safety requirements are fulfilled and that all safety-related parts of the software have been identified. In scope of this work product, safety measures shall be specified to mitigate potential failures related to ML. Additionally, it should include  the definition of verification criteria to confirm effectiveness of designed safety measures. \\ \hhline{|||~-} & \cellcolor{gray!25}DATASET: Incorrect road lane annotation could lead to decrease in final model performance, but impact of the quality of annotations is not considered in the current version of the safety concept.
	      \newline MODEL: A corrupted input image provided by a software communication layer to the DNN could cause a similar effect to that of an adversarial attack on the model. \\ \cline{1-2}

    \multirow{2}{*}{\linewrapcol{6-7.5.3 Dependent failures analysis report}} & Determine if the implementation of software safety requirements defined for ML components ensures independence and freedom from interference and how possible dependent failures can be mitigated. \\ \hhline{|||~-} & \cellcolor{gray!25}DATASET: It was identified that the second annotation set was prepared by transforming  the first annotation set through use of a constant offset, and so both annotations sets were most likely not prepared independently.
	    \newline MODEL: Two models use the same input camera sensor, and so corruption of that sensor might cause failure of both DNNs. Both algorithms were developed using the same ML framework, and thus the achieved the level of diversity is not sufficient (e.g., due to potential systematic fault in the implementation of a certain, commonly used DNN layer). \\ \cline{1-2}
    

    \multicolumn{2}{||c||}{Clause 6-8 Software unit design and implementation } \\ \cline{1-2} 
	    \multirow{2}{*}{\linewrapcol{6-8.5.1 Software unit design specification}} &  
	    Specify the functional behavior and the detailed information that is necessary for implementation of ML algorithms, KPIs, and runtime monitoring. It comprises unit design of the host and target frameworks (which are typically pre-existing). It should also include all internal models' parameters and hyperparameters and their optimization during the training and validation steps. This activity should be continued until the models achieve expected effectiveness. Finally, the output of this work product should be frozen models (i.e.,  deep neural networks including weights), which will not be changed or modified again.
   		\newline This work product  shall also cover the data acquisition and annotation processes. As was discussed for software architectural design, in our opinion, it is crucial to include dataset development to the V-model defined in ISO 26262 so as to define and implement integration of the data-collection process with model training. The output of this work product shall be a dataset with annotations divided into training, validation, and testing subsets. \\ \hhline{|||~-}
	    & \cellcolor{gray!25}DATASET: Road lane annotations shall be stored in separate XML files for each input image. The road lane specification shall define key feature points’ coordinates (x,y) to represent lanes, and each lane shall be stored in separate XML nodes.
	    \newline MODEL: Model training procedures shall be implemented using TensorFlow and PyTorch frameworks. The optimal parameters and hyperparameters defined for the LaneNet model shall be established during the training and validation process to achieve  the pre-defined  KPI values given the model robustness with respect to data augmentation, impact of different noise models, and adversarial attacks. 
    	\\ \cline{1-2}
    
	\multirow{2}{*}{\linewrapcol{6-8.5.2 Software unit implementation}}
	    & Implement the required ML algorithms (both host and target), perform training and validation of the selected models, optimize their internal parameters, and evaluate KPI and runtime monitoring metrics. In this part, the data shall be gathered and annotated. The results of this step shall be frozen ML models that are not to be changed during software unit testing. Obviously, until a subset of the annotated data is not provided, it is not possible to begin model training, but it could be realized as an iterative and parallel process while additional data samples are acquired and labeled \\ \hhline{|||~-}
		& \cellcolor{gray!25}DATASET: Gathering and annotation process of the images with road lanes.
        \newline MODEL: Implementation, training, and validation of the LaneNet network to detect the road lane according to the pre-defined KPI. \\ \cline{1-2}

    \multicolumn{2}{||c||}{Clause 6-9 Software unit verification } \\ \cline{1-2} 
	    \multirow{2}{*}{\linewrapcol{6-9.5.1 Software verification specification}} &
	    Specify how to test the final model on the test dataset according to the pre-defined KPI to achieve the expected performance. This work product shall also specify how the quality of the labeling process should be verified.  \\ \hhline{|||~-}
	    & \cellcolor{gray!25}DATASET: 20\% of the data shall be sent to other suppliers to verify if  all road lanes were correctly annotated. 
	     \newline MODEL: To evaluate the reproducibility of the model, the 10 independent LaneNet models shall be tested.
	     \newline NOTE: Values of the parameters in this example have been established arbitrarily for the purpose of the example, and those parameters shall be determined based on calculation of statistical significance.
	    \\ \hline

	\multirow{2}{*}{\linewrapcol{6-9.5.2 Software verification report}}
	    & Test frozen models (obtained during the training phase) on the test dataset according to the pre-defined  KPI, measuring the performance of implemented runtime monitors, and performing standard  verification testing recommended by ISO 26262. In the scope of this paper, we do not discuss which testing techniques are applicable to ML algorithms, but we would like to address this verification, which should be applied with respect to this work product on the models that were obtained following completion of the training process. An additional important part of this work product is also back-to-back testing between the training model and the runtime model. This work product shall also cover verification of the quality of the labeling process and whether quality of dataset labeling meets pre-defined  requirements. \\ \hhline{|||~-}
	    & \cellcolor{gray!25}DATASET: Verification as to whether road line boundaries were annotated pixel-by-pixel.
	        \newline MODEL:  Verification whether the reproducibility of the LaneNet model training has been confirmed. 
	         \\ \cline{1-2}

   \multicolumn{2}{||c||}{Clause 6-10 Software integration and verification } \\ \cline{1-2} 
	   \multirow{2}{*}{\linewrapcol{6-10.5.1 Software verification specification (refined)}} & Specify how to test the wider scope of the system, with both ML networks integrated, according to selected KPI and runtime monitoring metrics. This work product shall also specify how to verify the consistency between two subsets of annotations provided by independent annotators or suppliers. \\ \hhline{|||~-} & \cellcolor{gray!25}DATASET: Consistency between two independent road lane annotations obtained from different suppliers shall be evaluated by calculating inter-annotator variance.
        \newline MODEL: Performance of the two integrated road lane detectors—the LaneNet network and the Canny edge detector—shall be evaluated according to pre-defined  KPI. \\ \cline{1-2}
    
	   \multirow{2}{*}{\linewrapcol{6-10.5.2 Embedded software}}
	   & Define the steps that are required to integrate ML models implemented at software unit design level into larger ones until the embedded software is fully integrated. Given the dataset, in this work package, the steps to integrate the annotations that were generated from many annotators or suppliers should be defined. \\ \hhline{|||~-}
	   & \cellcolor{gray!25}DATASET: Consistency between the coordinates of road lane annotations generated by two annotators shall be analyzed in the first step. In the second step, consistency of the road lane annotations shall be evaluated between different suppliers.
	        \newline MODEL: In the first step, performance of two integrated road lane detectors shall be evaluated. In the second step, the road lane models shall be tested, together with runtime monitoring.  \\ \cline{1-2} \pagebreak

\cline{1-2} 
	   \multirow{2}{*}{\linewrapcol{6-10.5.3 Software verification report (refined)}}
	   & Test the performance of the integrated ML models according to the pre-defined  KPI and runtime monitors. Additionally, if the dataset annotations were prepared by independent annotators or suppliers, then this work product shall also provide a consistency verification report. Finally, this work product should reveal whether the consistency between dataset labeling annotations and integrated ML models is sufficient or if the software architectural design should be updated. \\ \hhline{|||~-}
	   &  \cellcolor{gray!25}DATASET: A report evaluating the consistency of two subsets of annotations provided by two independent suppliers.
	       \newline MODEL: A report summarizing performance of the integrated road lane detection models LaneNet and Canny edge detector. \\ \cline{1-2}
	    
	   \multicolumn{2}{||c||}{Clause 6-11 Testing of the embedded software } \\ \cline{1-2} 
	   \multirow{2}{*}{\linewrapcol{6-11.5.1 Software verification specification (refined)}}
	   & Specify the testing of the entire software including runtime monitoring on the target vehicle. This part shall also define the criteria to test whether a developed ML-based  component fulfills the software safety requirements of the target vehicle. Additionally, this part shall define how to test whether all software safety requirements defined for the dataset were fulfilled.  Finally, this work product shall specify how to test the integration of the whole process of data collection, labeling, and training of a new model in case runtime monitors detect edge cases that were incorrectly recognized by a currently developed ML component. \\ \hhline{|||~-}
	   & \cellcolor{gray!25}DATASET: Specifying  how to test if the dataset consists of the images gathered in winter conditions. 
	       \newline MODEL: Specifying  E2E testing of the road lane detection component. \\ \cline{1-2}
	    
	   \multirow{2}{*}{\linewrapcol{6-11.5.2 Software verification report (refined)}}
	    & According to the specification defined in the software verification specification, perform the final tests of an ML-based component and annotated dataset to finally confirm that they fulfill all allocated software safety requirements. \\ \hhline{|||~-}
	    & \cellcolor{gray!25}DATASET: The road lane detection dataset is incomplete, as it does not contain examples of highway roads recorded during winter conditions, and it shall be updated. 
	        \newline MODEL: The used runtime monitor designed for the road lane detection component is not able to identify winter conditions and so allows the highway autopilot to be turned on under winter conditions.
	    \\ \hline
	    
    \multicolumn{2}{|c|}{Part 7 Operation, service, and decommissioning} \\ \cline{1-2}     
 	  \multicolumn{2}{||c||}{Clause 7-7 Operation, service, and decommissioning} \\ \cline{1-2}     
		\multirow{2}{*}{\linewrapcol{7-7.5.1 Field observation instructions}} & Perform continuous monitoring of released vehicles after the start of production. A field monitoring process shall be capable detecting outdated ML-models, e.g., by identification of anomalies such as out-of-distribution samples, and detecting the reaction on pre-defined trigger events for (planned) evolutionary updates over the whole safety lifecycle. The field monitoring process shall especially be planned to determine what, when, by whom, and how often data shall be processed and updated. Field monitoring and change management shall be closely interconnected. \\ \hhline{|||~-}
   		 & \cellcolor{gray!25}Out-of-distribution detection running of a vehicle detects anomalous events that require a degradation of the ML-based function.	\\ \hline

    \multicolumn{2}{|c|}{Part 8 Supporting processes} \\ \cline{1-2} 
	\multicolumn{2}{||c||}{Clause 8-8 Change Management} \\ \cline{1-2}     
		\multirow{2}{*}{\linewrapcol{8-8.5.1 Change management plan}}
	    & Specify a change management plan that applies to a project that is under continuous delivery (as opposed to one with just one delivery—start of production). ML-based systems will very likely have to be updated continuously over the product lifetime. Change management planning therefore shall consider all foreseeable trigger events that could possibly imply a change, such as explicitly planned continuous changes, changes necessary due to detected anomalies, or changes due to aging of demands. This analysis and planning activity shall be performed at an early stage of development. Having mature change management planning available will be essential to automating activities by incorporating the backend tool chain. \\ \hhline{|||~-}
	    & \cellcolor{gray!25}Aging of demands: Due to a change in legislation, new test cases are mandatory, leading to an update (re-training) of the ML-based system.
	    \\ \cline{1-2}
    
   \multicolumn{2}{||c||}{Clause 8-11 Confidence in the use of software tools } \\ \cline{1-2} 
	\multirow{2}{*}{\linewrapcol{8-11.5.1 Software tool criteria evaluation report}} & Evaluate the impact of the ML tools (i.e., framework for implementation of DNN, training and inference procedures, ML lifecycle support tools) that are used during the development process and deployment of the released product. Errors that they introduce (e.g., generators, training) or errors that they detect (ML simulation or testing tools) are hard to be detected by other means, and so there is a potential risk of having several complex tools having a tool confidence level  of 2 or 3. As a result, such tools would need to be qualified for safety according to the allocated ASIL, a possibly troublesome task due to these tools’ high levels of complexity. \\ \hhline{|||~-} &  \cellcolor{gray!25}Diversity in model training implementation is provided by the usage of two development frameworks (PyTorch and TensorFlow), both frameworks have  different internal architectures and the final models are extensively tested. Therefore, the frameworks are classified as TCL1. \\ \cline{1-2}
  
    \multirow{2}{*}{\linewrapcol{8-11.5.2 Software tool qualification report}} & 	Perform a tool evaluation and, if needed, safety-qualification of ML tools. \\ \hhline{|||~-} & \cellcolor{gray!25}A selected TCL2 framework to manage the ML lifecycle has a well-defined, scrum-based development process; the software is deployed in thousands of hosts in the evaluated major version; and there are extensive unit tests for the tool as well as an independently developed validation test framework covering all safety-related features of the tool. All of these allow this tool to achieve TCL2. \\
    \hline

\end{longtable}


\endgroup

\endgroup

\begin{multicols}{2}
\section{Conclusions}
In this paper, we propose how ML-based products can be developed as currently specified according to the ISO 26262 and ISO/PAS 21448 standards. Our work described herein has been motivated primarily by the lack of a development process definition compatible with ISO 26262 with respect to ML-based products in road vehicles.

\par With this work, we also seek to show that design and development of ML-based components can be organized using the existing version of the ISO 26262 standard, after some adaptations have been made. Defining the whole process according to the workflow proposed by ISO 26262 facilitates tracing potential failures and limitations in the design and development process of autonomous driving that utilizes ML-based products. The proposed organization of product development can be applied immediately, even before a dedicated automotive standard will be released.
We believe that our work will contribute to further progress in applying ML algorithms within the automotive industry.

\section*{Acknowledgment}
\addcontentsline{toc}{section}{Acknowledgment}
	The authors would like to thank  Dr. Jelena Frtunikj for providing constructive recommendations and remarks which helped to improve and clarify this manuscript.
\end{multicols}


\begin{multicols}{2}
\bibliographystyle{IEEEtran}
\balance
\bibliography{bibliography} 

\begin{thebibliography}{10}
\providecommand{\url}[1]{#1}
\csname url@samestyle\endcsname
\providecommand{\newblock}{\relax}
\providecommand{\bibinfo}[2]{#2}
\providecommand{\BIBentrySTDinterwordspacing}{\spaceskip=0pt\relax}
\providecommand{\BIBentryALTinterwordstretchfactor}{4}
\providecommand{\BIBentryALTinterwordspacing}{\spaceskip=\fontdimen2\font plus
\BIBentryALTinterwordstretchfactor\fontdimen3\font minus
  \fontdimen4\font\relax}
\providecommand{\BIBforeignlanguage}[2]{{%
\expandafter\ifx\csname l@#1\endcsname\relax
\typeout{** WARNING: IEEEtran.bst: No hyphenation pattern has been}%
\typeout{** loaded for the language `#1'. Using the pattern for}%
\typeout{** the default language instead.}%
\else
\language=\csname l@#1\endcsname
\fi
#2}}
\providecommand{\BIBdecl}{\relax}
\BIBdecl

\bibitem{YoungHazarika2018}
T.~{Young}, D.~{Hazarika}, S.~{Poria}, and E.~{Cambria}, ``Recent trends in
  deep learning based natural language processing [review article],''
  \emph{IEEE Computational Intelligence Magazine}, vol.~13, no.~3, pp. 55--75,
  2018.

\bibitem{NassifShahin2019}
A.~B. {Nassif}, I.~{Shahin}, I.~{Attili}, M.~{Azzeh}, and K.~{Shaalan},
  ``Speech recognition using deep neural networks: A systematic review,''
  \emph{IEEE Access}, vol.~7, pp. 19\,143--19\,165, 2019.

\bibitem{IngleFujii2019}
R.~R. {Ingle}, Y.~{Fujii}, T.~{Deselaers}, J.~{Baccash}, and A.~C. {Popat}, ``A
  scalable handwritten text recognition system,'' in \emph{2019 International
  Conference on Document Analysis and Recognition (ICDAR)}, 2019, pp. 17--24.

\bibitem{LiuWand2017}
W.~Liu, Z.~Wang, X.~Liu, N.~Zeng, Y.~Liu, and F.~E. Alsaadi, ``A survey of deep
  neural network architectures and their applications,'' \emph{Neurocomputing},
  vol. 234, pp. 11 -- 26, 2017.

\bibitem{da2019recommendation}
A.~Da’u and N.~Salim, ``Recommendation system based on deep learning methods:
  a systematic review and new directions,'' \emph{Artificial Intelligence
  Review}, pp. 1--40, 2019.

\bibitem{GrigorescuTrasnea2020}
S.~Grigorescu, B.~Trasnea, T.~Cocias, and G.~Macesanu, ``A survey of deep
  learning techniques for autonomous driving,'' \emph{Journal of Field
  Robotics}, vol.~37, no.~3, pp. 362--386, 2020.

\bibitem{UL4600}
\BIBentryALTinterwordspacing
UL, ``{UL 4600}, standard for evaluation of autonomous products, {E}dition~1.''
  [Online]. Available: \url{https://ul.org/UL4600}
\BIBentrySTDinterwordspacing

\bibitem{ISO-26262-2.0}
\BIBentryALTinterwordspacing
ISO, ``{ISO 26262} ({P}art 1-12) -- {R}oad {V}ehicles -- {F}unctional {S}afety,
  {S}econd edition,'' December 2018. [Online]. Available:
  \url{http://www.iso.org}
\BIBentrySTDinterwordspacing

\bibitem{ISO-21448}
\BIBentryALTinterwordspacing
------, ``{ISO/PAS} 21448 -- {R}oad {V}ehicles - {S}afety of the intended
  functionality, first edition,'' January 2019. [Online]. Available:
  \url{https://www.iso.org/standard/70939.html}
\BIBentrySTDinterwordspacing

\bibitem{IEC-61508}
IEC, ``{{IEC} 61508 Part 1-7 Second Edition},'' Geneva, 2010.

\bibitem{SalayQueiroz2018}
R.~Salay, R.~Queiroz, and K.~Czarnecki, ``An analysis of {ISO} 26262: Machine
  learning and safety in automotive software,'' in \emph{SAE Technical
  Paper}.\hskip 1em plus 0.5em minus 0.4em\relax SAE International, 04 2018.

\bibitem{KuwajimaYasuoka2020}
H.~Kuwajima, H.~Yasuoka, and T.~Nakae, ``Engineering problems in machine
  learning systems,'' \emph{Machine Learning}, vol. 109, p. 1103–1126, 2020.

\bibitem{SchwalbeSchels2020}
G.~Schwalbe and M.~Schels, ``{A Survey on Methods for the Safety Assurance of
  Machine Learning Based Systems},'' in \emph{{10th European Congress on
  Embedded Real Time Software and Systems (ERTS 2020)}}, Toulouse, France,
  2020.

\bibitem{huang2018survey}
X.~Huang, D.~Kroening, W.~Ruan, J.~Sharp, Y.~Sun, E.~Thamo, M.~Wu, and X.~Yi,
  ``A survey of safety and trustworthiness of deep neural networks,'' 2018.

\bibitem{Geiger2012CVPR}
A.~Geiger, P.~Lenz, and R.~Urtasun, ``Are we ready for autonomous driving? the
  kitti vision benchmark suite,'' in \emph{Conference on Computer Vision and
  Pattern Recognition (CVPR)}, 2012.

\bibitem{Fritsch2013ITSC}
J.~Fritsch, T.~Kuehnl, and A.~Geiger, ``A new performance measure and
  evaluation benchmark for road detection algorithms,'' in \emph{International
  Conference on Intelligent Transportation Systems (ITSC)}, 2013.

\bibitem{ZendelMurschitz2015}
O.~{Zendel}, M.~{Murschitz}, M.~{Humenberger}, and W.~{Herzner}, ``Cv-hazop:
  Introducing test data validation for computer vision,'' in \emph{2015 IEEE
  Int. Conf. on Computer Vision (ICCV)}, 2015, pp. 2066--2074.

\bibitem{SAEJ3016}
\BIBentryALTinterwordspacing
SAE, ``Taxonomy and definitions for terms related to on-road motor vehicle
  automated driving systems.'' [Online]. Available:
  \url{https://www.sae.org/standards/content/j3016\_201401/}
\BIBentrySTDinterwordspacing

\bibitem{ISO-21434}
\BIBentryALTinterwordspacing
ISO, ``{ISO/SAE} 21434 -- {R}oad {V}ehicles – {C}ybersecurity {E}ngineering,
  (under development).'' [Online]. Available:
  \url{https://www.iso.org/standard/70918.html}
\BIBentrySTDinterwordspacing

\bibitem{KoopmanFerrell2019}
P.~Koopman, U.~Ferrell, F.~Fratrik, and M.~Wagner, ``A safety standard approach
  for fully autonomous vehicles,'' in \emph{Computer Safety, Reliability, and
  Security}, A.~Romanovsky, E.~Troubitsyna, I.~Gashi, E.~Schoitsch, and
  F.~Bitsch, Eds., 2019, pp. 326--332.

\bibitem{safad2019}
\BIBentryALTinterwordspacing
Aptiv, Audi, Baidu, BMW, Continental, Daimler, {FCA US LLC}, HERE, Infineon,
  Intel, and Volkswagen, ``{Safety First for Automated Driving},'' Tech. Rep.,
  July 2019. [Online]. Available:
  \url{https://www.press.bmwgroup.com/global/article/attachment/T0298103EN/434404}
\BIBentrySTDinterwordspacing

\bibitem{VDE2842}
\BIBentryALTinterwordspacing
VDE, ``{VDE-AR-E} 2842-61-2 --development and trustworthiness of
  autonomous/cognitive systems – part 61-2: Management.'' [Online].
  Available:
  \url{https://www.vde-verlag.de/standards/1800575/e-vde-ar-e-2842-61-2-anwendungsregel-2020-07.html}
\BIBentrySTDinterwordspacing

\bibitem{ISOSC42}
\BIBentryALTinterwordspacing
ISO, ``{ISO/IEC JTC 1/SC 42} {A}rtificial intelligence.'' [Online]. Available:
  \url{https://www.iso.org/committee/6794475.html}
\BIBentrySTDinterwordspacing

\bibitem{Rudzicz2019}
F.~Rudzicz, P.~Paprica, and M.~Janczarski, ``Towards international standards
  for evaluating machine learning,'' in \emph{SafeAI at AAAI19}, 2019.

\bibitem{ISO-4804}
\BIBentryALTinterwordspacing
ISO, ``{ISO/CD TR} 4804 -- {R}oad {V}ehicles – {S}afety and cybersecurity for
  automated driving systems — {D}esign, verification and validation methods,
  (under development).'' [Online]. Available:
  \url{https://www.iso.org/standard/80363.html}
\BIBentrySTDinterwordspacing

\bibitem{SalayCzarnecki2018}
\BIBentryALTinterwordspacing
R.~Salay and K.~Czarnecki, ``Using machine learning safely in automotive
  software: An assessment and adaption of software process requirements in
  {ISO} 26262,'' vol. abs/1808.01614, 2018. [Online]. Available:
  \url{http://arxiv.org/abs/1808.01614}
\BIBentrySTDinterwordspacing

\bibitem{HenrikssonBorg2018}
J.~{Henriksson}, M.~{Borg}, and C.~{Englund}, ``Automotive safety and machine
  learning: Initial results from a study on how to adapt the {ISO} 26262 safety
  standard,'' in \emph{2018 IEEE/ACM 1st International Workshop on Software
  Engineering for AI in Autonomous Systems (SEFAIAS)}, 2018, pp. 47--49.

\bibitem{Spanfelner2012}
B.~Spanfelner, D.~Richter, S.~Ebel, U.~Wilhelm, W.~Branz, and C.~Patz.,
  ``Challenges in applying the {ISO} 26262 for driver assistance systems,''
  \emph{Tagung Fahrerassistenz}, 2012.

\bibitem{SalayCzrnecki2019}
R.~Salay and K.~Czarnecki, ``Improving ml safety with partial specifications,''
  in \emph{Computer Safety, Reliability, and Security}, A.~Romanovsky,
  E.~Troubitsyna, I.~Gashi, E.~Schoitsch, and F.~Bitsch, Eds.\hskip 1em plus
  0.5em minus 0.4em\relax Cham: Springer International Publishing, 2019, pp.
  288--300.

\bibitem{MohseniPitale2020}
S.~Mohseni, M.~Pitale, V.~Singh, and Z.~Wang, ``Practical solutions for machine
  learning safety in autonomous vehicles,'' in \emph{Proc. of the Workshop on
  Artificial Intelligence Safety, co-located with 34th {AAAI} Conference on
  Artificial Intelligence}, ser. {CEUR} Workshop Proceedings, vol. 2560.\hskip
  1em plus 0.5em minus 0.4em\relax CEUR-WS.org, 2020, pp. 162--169.

\bibitem{WillersSudholt2020}
\BIBentryALTinterwordspacing
O.~Willers, S.~Sudholt, S.~Raafatnia, and S.~Abrecht, ``Safety concerns and
  mitigation approaches regarding the use of deep learning in safety-critical
  perception tasks,'' \emph{CoRR}, vol. abs/2001.08001, 2020. [Online].
  Available: \url{https://arxiv.org/abs/2001.08001}
\BIBentrySTDinterwordspacing

\bibitem{ChengHuang2018}
C.~{Cheng}, C.~{Huang}, and G.~{Nührenberg}, ``nn-dependability-kit:
  Engineering neural networks for safety-critical autonomous driving systems,''
  in \emph{2019 IEEE/ACM International Conference on Computer-Aided Design
  (ICCAD)}, 2019, pp. 1--6.

\bibitem{SzeChen2017}
V.~{Sze}, Y.~{Chen}, T.~{Yang}, and J.~S. {Emer}, ``Efficient processing of
  deep neural networks: A tutorial and survey,'' \emph{Proceedings of the
  IEEE}, vol. 105, no.~12, pp. 2295--2329, 2017.

\bibitem{BadueGuidolini2019}
C.~Badue, R.~Guidolini, R.~V. Carneiro, P.~Azevedo, V.~B. Cardoso, A.~Forechi,
  L.~F.~R. Jesus, R.~F. Berriel, T.~M. Paixão, F.~Mutz, T.~Oliveira-Santos,
  and A.~F.~D. Souza, ``Self-driving cars: A survey,'' 2019.

\bibitem{SenguptaBasak2019}
S.~Sengupta, S.~Basak, P.~Saikia, S.~Paul, V.~Tsalavoutis, F.~Atiah, V.~Ravi,
  and A.~Peters, ``A review of deep learning with special emphasis on
  architectures, applications and recent trends,'' 2019.

\bibitem{bojarski2016end}
M.~Bojarski, D.~D. Testa, D.~Dworakowski, B.~Firner, B.~Flepp, P.~Goyal, L.~D.
  Jackel, M.~Monfort, U.~Muller, J.~Zhang, X.~Zhang, J.~Zhao, and K.~Zieba,
  ``End to end learning for self-driving cars,'' 2016.

\bibitem{Zendel2017}
O.~{Zendel}, K.~{Honauer}, M.~{Murschitz}, M.~{Humenberger}, and G.~F.
  {Domínguez}, ``Analyzing computer vision data — the good, the bad and the
  ugly,'' in \emph{2017 IEEE Conference on Computer Vision and Pattern
  Recognition (CVPR)}, 2017, pp. 6670--6680.

\bibitem{feng2019deep}
D.~Feng, C.~Haase-Schuetz, L.~Rosenbaum, H.~Hertlein, C.~Glaeser, F.~Timm,
  W.~Wiesbeck, and K.~Dietmayer, ``Deep multi-modal object detection and
  semantic segmentation for autonomous driving: Datasets, methods, and
  challenges,'' 2019.

\bibitem{SatzodaEvaluationMetrics}
R.~Satzoda and M.~Trivedi, ``On performance evaluation metrics for lane
  estimation,'' \emph{Proceedings - International Conference on Pattern
  Recognition}, pp. 2625--2630, 12 2014.

\bibitem{ChengNuhrenberg2019}
C.~{Cheng}, G.~{Nührenberg}, and H.~{Yasuoka}, ``Runtime monitoring neuron
  activation patterns,'' in \emph{2019 Design, Automation Test in Europe
  Conference Exhibition (DATE)}, 2019, pp. 300--303.

\bibitem{Kendall2017}
A.~Kendall and Y.~Gal, ``What uncertainties do we need in bayesian deep
  learning for computer vision?'' in \emph{Proceedings of the 31st Int. Conf.
  on Neural Information Processing Systems}, ser. NIPS'17, 2017, pp.
  5580--5590.

\bibitem{clemensalex2018active}
C.-A. Brust, C.~Käding, and J.~Denzler, ``Active learning for deep object
  detection,'' \emph{Proceedings of the 14th International Joint Conference on
  Computer Vision, Imaging and Computer Graphics Theory and Applications},
  2019.

\bibitem{Roy2018}
S.~Roy, A.~Unmesh, and V.~P. Namboodiri, ``Deep active learning for object
  detection,'' in \emph{British Machine Vision Conference}, 2018.

\bibitem{wang2018lanenet}
Z.~Wang, W.~Ren, and Q.~Qiu, ``Lane{N}et: Real-time lane detection networks for
  autonomous driving,'' 2018.

\bibitem{WuChang2014}
P.-C. Wu, C.-Y. Chang, and C.~H. Lin, ``Lane-mark extraction for automobiles
  under complex conditions,'' \emph{Pattern Recognition}, vol.~47, no.~8, pp.
  2756 -- 2767, 2014.

\end{thebibliography}
\balance
\end{multicols}
\clearpage



\end{document}